\documentclass[10pt,twocolumn,letterpaper]{article}

\usepackage[pagenumbers]{cvpr} % To force page numbers, e.g. for an arXiv version

\usepackage{graphicx}
\usepackage{amsmath}
\usepackage{amssymb}
\usepackage{bbm}
\usepackage{booktabs}
\usepackage{cite}
\usepackage{bbding}

\usepackage{multirow}
\usepackage{algorithm}
\usepackage{algorithmicx}
\usepackage{algpseudocode}
\usepackage{amsthm,amsmath,amssymb}
\usepackage{subcaption}	

\usepackage{xcolor}
\newcommand{\mypm}[1]{\color{gray}{\tiny{#1}}}

\usepackage[pagebackref=true,breaklinks=true,colorlinks,bookmarks=false]{hyperref}

\floatname{algorithm}{Algorithm}

\usepackage{cleveref}
\crefname{fig}{Fig.}{Fig.}
\crefname{subfig}{Fig.}{Fig.}
\crefname{table}{Table}{Table}
\crefname{equation}{Eq.}{Eq.}
\crefname{Algorithm}{Algorithm}{Algorithm}
\crefname{subsec}{Sec}{Sec}

\def\cvprPaperID{540} % *** Enter the CVPR Paper ID here
\def\confName{CVPR}
\def\confYear{2022}

\begin{document}
\title{Active Teacher for Semi-Supervised Object Detection}
%%%%%%%%%%%%%% TITLE - PLEASE UPDATE %%%%%%%%%%%%%%
\author{
	Peng Mi$^{1}$\footnotemark[1],
	Jianghang Lin$^{1}$\footnotemark[1],
% 	\thanks{Equal Contribution. $\dagger$ Corresponding Author.},
	Yiyi Zhou$^{1}$\footnotemark[1]\thanks{Equal Contribution. $\dagger$ Corresponding Author.},
	Yunhang Shen$^{1}$,
	Gen Luo$^{1}$,
	Xiaoshuai Sun$^{1}$,
	\\
	Liujuan Cao$^{1\dagger}$,
	Rongrong Fu$^{2}$,
	Qiang Xu$^{2}$,
	Rongrong Ji$^{1}$
	 \\
	$^1$Media Analytics and Computing Lab,  School of Informatics, Xiamen University, 361005, China.\\
	$^2$Ascend Enabling Laboratory, Huawei Technologies, China.\\
	{\tt\small \{mipeng,hunterjlin007,luogen\}@stu.xmu.edu.cn}, {\tt\small \{zhouyiyi,xssun,caoliujuan,rrji\}@xmu.edu.cn},\\
% 	{\tt\small odysseyshen@tencent.com},
    {\tt\small shenyunhang01@gmail.com},
	{\tt\small \{furongrong, xuqiang40\}@huawei.com}
}

\maketitle

%%%%%%%%%%%%%%%%%% Abstract %%%%%%%%%%%%%%%%%%
\begin{abstract}
In this paper, we study teacher-student learning from the perspective of data initialization and propose a novel algorithm called Active Teacher\footnote{Source code are available at: \url{https://github.com/HunterJ-Lin/ActiveTeacher}} for semi-supervised object detection (SSOD). Active Teacher extends the teacher-student framework to an iterative version, where the label set is partially initialized and gradually augmented by evaluating three key factors of unlabeled examples, including difficulty, information and diversity. With this design, Active Teacher can maximize the effect of limited label information while improving the quality of pseudo-labels. To validate our approach, we conduct extensive experiments on the MS-COCO benchmark and compare Active Teacher with a set of recently proposed SSOD methods. The experimental results not only validate the superior performance gain of Active Teacher over the compared methods, but also show that it enables the baseline network, \ie, Faster-RCNN, to achieve $100$\% supervised performance with much less label expenditure, \ie $40$\% labeled examples on MS-COCO. More importantly, we believe that the experimental analyses in this paper can provide useful empirical knowledge for data annotation in practical applications.
\end{abstract}

% \vspace{-2mm}
%%%%%%%%%%%%%%%%%% Intro %%%%%%%%%%%%%%%%%%
\section{Introduction}
\label[section]{sec:intro}

Recent years have witnessed the rapid development of object detection supported by a flurry of benchmark datasets~\cite{voc, coco, object365, imagenet} and methods~\cite{rcnn, fastrcnn, fasterrcnn, yolo, yolov3, yolox, ssd}. Despite great success, the expensive instance-level annotation has long plagued the advancement and application of existing detection models. To this end, how to save labeling expenditure has become a research focus in object detection~\cite{weak-wsddn, weak-oicr, weak-uwsod, misra2015watchandlearn-ssod, consistency-based-ssod, SSM-active-ssod, tang2016largescalesemi-ssod, collaborative-ssod, stac-ssod-ts, unbiasedteacher-ssod-ts, consistency-based-ssod, yolov4}.

\begin{figure}[t]
    \centering
    \includegraphics[scale=0.51]{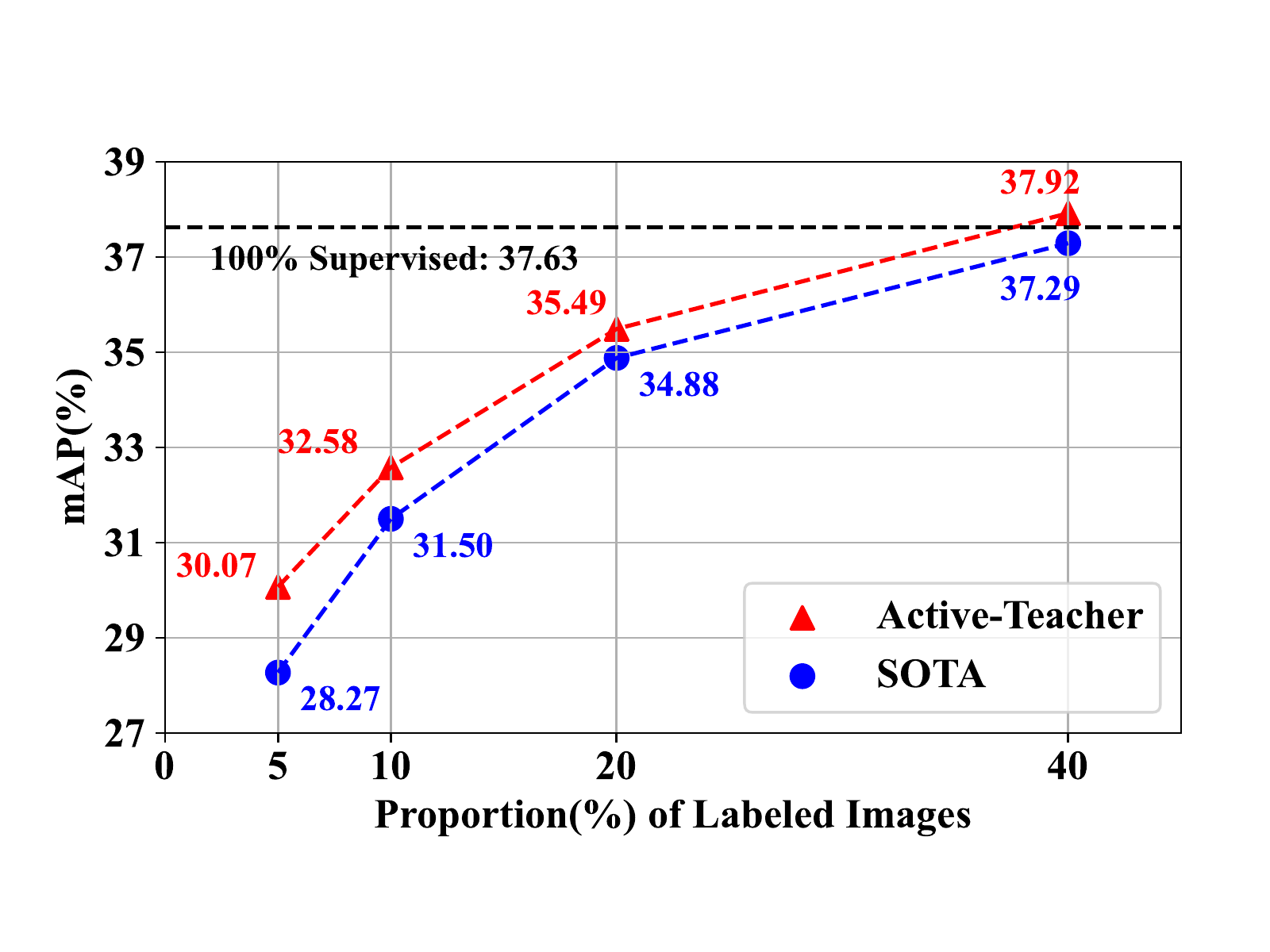}
    \vspace{-10mm}
    \caption{The performance comparison between Active Teacher and the state-of-the-art~(SOTA) method\cite{unbiasedteacher-ssod-ts} with different proportions of labeled data in MS-COCO. Active Teacher exceeds $100\%$ fully supervised performance with only $40$\% label information.}
    \vspace{-4mm}
    \label[fig]{fig:percent2supervision}
\end{figure}

Inspired by recent success in image classification~\cite{temporal-ssl, meanteacher-ssl-ts, fixmatch-ssl, mixmatch-ssl, remixmatch-ssl}, some practitioners resort to teacher-student learning for semi-supervised object detection~(SSOD)~\cite{tang2021humbleteacher-ssod-ts, stac-ssod-ts, unbiasedteacher-ssod-ts}. Specifically, this methodology uses a teacher network with weakly augmented labeled data to generate high-quality pseudo-labels for the student network with strong data augmentation~\cite{cutout, autoaugment, dataaug_forconsistency-ssl}. This self-training process helps the model explore large amounts of unlabeled data based on a very limited number of annotations. Following this methodology, Sohn~\etal~\cite{stac-ssod-ts} proposed the first teacher-student framework called STAC for SSOD. This simple framework outperforms the existing semi-supervised methods~\cite{tang2016largescalesemi-ssod, weak-oicr, weak-uwsod, weak-wsddn} by a large margin, showing the great potential of teacher-student learning in object detection. 

%Based on STAC, some very recent works are proposed to address the shortcomings of teacher-student learning in SSOD~\cite{unbiasedteacher-ssod-ts, instant-ssod-ts, tang2021humbleteacher-ssod-ts}.
Some very recent SSOD works~\cite{unbiasedteacher-ssod-ts, instant-ssod-ts, tang2021humbleteacher-ssod-ts} are proposed to improve this methodology.
For instances, Liu~\etal~\cite{unbiasedteacher-ssod-ts} apply \emph{exponential moving average}~(EMA)~\cite{meanteacher-ssl-ts} to train a gradually progressing teacher to alleviate the class-imbalance and over-fitting issues. Zhou~\etal~\cite{instant-ssod-ts} propose an instant pseudo labeling strategy to reduce the impact of the confirmation bias and improve the quality of pseudo labeling. In~\cite{tang2021humbleteacher-ssod-ts}, Tang~\etal adopt a detection-specific data ensemble to produce more reliable pseudo-labels.
%Conclusively, these methods mainly focus on the optimization of the framework or reducing the negative impacts of pseudo labels, of which contributions are orthogonal to our work.
Conclusively, these methods mainly focus on the framework optimization or the negative impact of noisy pseudo-labels, of which contributions are orthogonal to ours.

In this paper, we study this semi-supervised methodology from the perspective of data initialization. More specifically, we investigate how to select the optimal labeled examples for teacher-student learning in SSOD. To explain, although a plenty of pseudo-labels are generated for self-training, ground-truth label information still plays a key role in the infant training phase, which determines the quality of pseudo-labels and the performance lower-bound of the teacher networks~\cite{stac-ssod-ts, unbiasedteacher-ssod-ts, instant-ssod-ts}.  Meanwhile, in some teacher-student methods\cite{unbiasedteacher-ssod-ts, roychowdhury2019automatic-ssod}, the pseudo-labels are only used to optimize the predictions of object categories and foreground-background proposals, while the optimization of bounding boxes regression still relies on the ground-truth annotations. In this case, we observe that ground-truth label information plays an important role in SSOD, which, however, is still left unexplored.

\begin{figure*}[t]
    \centering
    \includegraphics[scale=0.6,trim=90 130 45 90,clip]{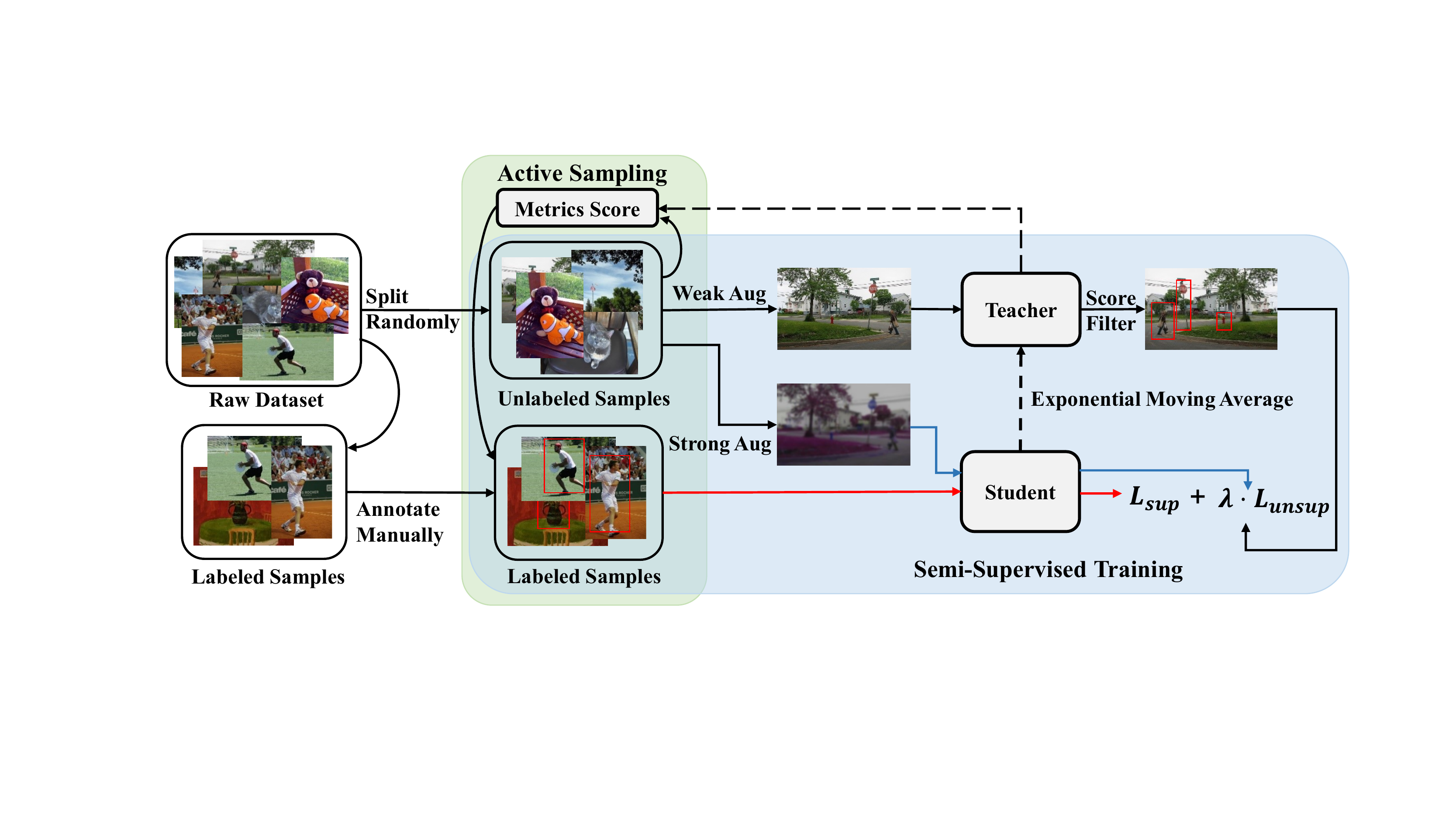}
    % \vspace{-2mm}
    \caption{The overall framework of the proposed Active Teacher. In Active Teacher, the label set is partially initialized and gradually augmented after each semi-supervised training. Active Teacher includes two detection networks, \ie, Faster-RCNN~\cite{fasterrcnn}, with the same configurations, namely \emph{Teacher} and \emph{Student}. Teacher is used to generate pseudo-labels for training Student, and its parameters are gradually updated from Student via EMA\cite{meanteacher-ssl-ts}. Student is trained with both ground-truth and pseudo-labels, denoted as $\mathcal{L}_{sup}$ and $\mathcal{L}_{unsup}$, respectively. Teacher also serves to estimate the unlabeled examples for active sampling.}
    % \vspace{-2mm}
    \label[fig]{fig:framework}
\end{figure*}

To this end, we propose a new teacher-student method, coined as \emph{Active Teacher}, for semi-supervised object detection. As shown in~\cref{fig:framework}, Active Teacher extends the conventional teacher-student framework to an iterative one, where the label set is partially initialized and gradually augmented via a novel active sampling strategy. With this modification, Active Teacher can maximize the effect of limited label information by active sampling, which can also improve the quality of pseudo-labels. We further investigate the selection of labeled examples from the aspects of \emph{difficulty}, \emph{information} and \emph{diversity}, and the values of these metrics are automatically combined without hyper-parameter tunning. Through these metrics, we can explore what kind of data are optimal for SSOD.

To validate the proposed method, we conduct extensive experiments on the benchmark dataset, namely MS COCO~\cite{coco}\footnote{More experimental results can be found in our Github project.}. The experimental results not only confirm the significant performance gains of Active Teacher against a set of state-of-the-art SSOD methods, \eg, $+6.3\%$ and $+23.3\%$ compared with Unbiased Teacher~\cite{unbiasedteacher-ssod-ts} and STAC~\cite{stac-ssod-ts} on $5\%$ MS-COCO, respectively. It also shows that Active Teacher enables the baseline detection network, \ie, Faster-RCNN~\cite{fasterrcnn}, to achieve $100\%$ supervised performance with much less labeling expenditure, \eg, with $40\%$ labeled examples on MS-COCO, as shown in~\cref{fig:percent2supervision}. More importantly, we also provide the in-depth analyses for active sampling, which can give useful hints for data annotation in practical applications of object detection. 

In summary, our contribution is two-fold:
\begin{itemize}
\item We present the first attempt of studying data initialization in teacher-student based semi-supervised object detection~(SSOD), and conduct extensive experiments for different sampling strategies. These quantitative and qualitative analyses can provide useful references for data annotation in practical applications.

\item We propose a new teacher-student framework for SSOD called \textit{Active Teacher}, which not only outperforms a set of SSOD methods on the benchmark dataset, but also enables the baseline detection network achieve 100\% fully supervised performance with much less label expenditure. 
\end{itemize}

%%%%%%%%%%%%%%%%%% Related Work %%%%%%%%%%%%%%%%%%
\section{Related Work}
\textbf{Object Detection.} With the rapid development of deep neural networks, object detection has achieved great progress  both academically and industrially~\cite{yolo, yolo9000, yolov3, ssd, rcnn, fastrcnn, fasterrcnn, maskrcnn, fpn, yolox}. Object detection is roughly divided into two genres: one-stage and two-stage detectors. The representative work of one-stage methods includes YOLO~\cite{yolo, yolo9000, yolov3,yolox}, SSD~\cite{ssd}, \etc, and the ones of two-stage models include RCNN series~\cite{rcnn, fastrcnn, fasterrcnn} and its variants~\cite{maskrcnn, fpn}. The main difference between these two methodologies is that the one-stage method directly predicts the coordinates and probability distribution of the object based on the feature map, while the two-stage methods use region proposal networks~\cite{fasterrcnn} to sample potential objects, and further predict the probability distribution and coordinate information of the object, respectively. Following the prior works~\cite{unbiasedteacher-ssod-ts, stac-ssod-ts, instant-ssod-ts}, we focus the semi-supervised learning of two-stage models and use Faster-RCNN~\cite{fasterrcnn} as our baseline network.

%, trim=left down right top,clip

\textbf{Semi-Supervised Object Detection.} In the field of computer vision, most existing researches on semi-supervised learning mainly focus on image classification~\cite{temporal-ssl, 5g, learning-with-pseudo-ensembles-ssl, self-training-nosiy-ssl-ts}, which can be roughly divided into consistency-based and pseudo labeling based methods, respectively . Consistency-based approaches~\cite{consistency-based-ssod, consistency-ssl, fixmatch-ssl, mixmatch-ssl, remixmatch-ssl} constrain the model to make it robust to noise via producing consistent prediction results. 
%Pseudo labeling~\cite{fixmatch-ssl, learning-with-pseudo-ensembles-ssl, tang2021humbleteacher-ssod-ts, stac-ssod-ts, unbiasedteacher-ssod-ts, instant-ssod-ts} methods firstly use labeled data to train a model and then use this model to generate pseudo-labels, and finally retrain model with labeled data and unlabeled data with pseudo-labels.
Pseudo labeling~\cite{fixmatch-ssl, learning-with-pseudo-ensembles-ssl, tang2021humbleteacher-ssod-ts, stac-ssod-ts, unbiasedteacher-ssod-ts, instant-ssod-ts} methods firstly train the classifiers with ground-truth annotations and generate pseudo-labels for unlabeled data, and finally retrain models with all data.
Recently, some works~\cite{consistency-based-ssod, jeong2021interpolation-ssod, stac-ssod-ts, instant-ssod-ts, unbiasedteacher-ssod-ts} apply semi-supervised learning to object detection. CSD~\cite{consistency-based-ssod} randomly flips images multiple times, driving the model to produce consistent predictions for these flipped images. ISD~\cite{jeong2021interpolation-ssod} uses \emph{mixup}~\cite{zhang2017mixup} to constrain model training. Following the popular teacher-student framework~\cite{meanteacher-ssl-ts}, STAC~\cite{stac-ssod-ts} proposes the first teacher-student based framework for SSOD. Due to the static annotating strategy, the pseudo-labels in STAC are fixed, which limits the final detection performance. In Instant-Teaching~\cite{instant-ssod-ts}, both teacher and student share the same parameters to deal with above problem. However, they still suffer from extreme instability in the initial training phase and require a high confidence score threshold for generating pseudo-labels. Unbiased teacher~\cite{unbiasedteacher-ssod-ts} exploits EMA~\cite{meanteacher-ssl-ts} to optimize teacher from student gradually.
%In addition, the focal loss\cite{lin2017focalloss} is used in~\cite{unbiasedteacher-ssod-ts} and~\cite{unbiasedteacher-ssod-ts} gets rid of the regression loss of unlabeled data.
In addition, Unbiased teacher apply EMA \cite{meanteacher-ssl-ts} and focal loss \cite{lin2017focalloss} to address the pseudo-label over-fitting problem in teacher-student learning. 

\textbf{Active Learning.} Efficient learning is widely studied\cite{zhang2021you, zheng2023ddpnas, zhang2023targeted}. There are also some active-learning based methods proposed to reduce the labeling expenditure of object detection~\cite{DGM-ac, cald-ac, MI-AOD-ac}. For instance, Wang \etal \cite{DGM-ac} use different active sampling metrics for different stages in object detection. CALD~\cite{cald-ac} measures information by calculating the data consistency of bounding boxes before and after augmentation. MI-AOD~\cite{MI-AOD-ac} applies multi-instance learning to suppress the  pseudo-label noises.

In this paper, we focus on the teacher-student based semi-supervised learning for object detection.

\vspace{-3mm}
%%%%%%%%%%%%%%%%%% Method %%%%%%%%%%%%%%%%%%
\section{Active Teacher}

The overall framework of the proposed Active Teacher is illustrated in \cref{fig:framework}. As shown in this figure, Active Teacher consists of an iterative teacher-student structure, where the limited label set is partially initialized and gradually augmented. After each iteration, the well-trained teacher network is used to evaluate the importance of unlabeled examples in terms of the proposed metrics, \ie, \emph{information}, \emph{diversity} and \emph{difficulty}, based on which active data augmentation is performed. The detailed procedure is depicted in \cref{alg:pseudocode_of_active_teacher}. In the following section, we introduce Active Teacher from the aspects of semi-supervised learning and active sampling, respectively. 

\vspace{-2mm}
% \vspace{+3mm}
\begin{algorithm}[t]
% \vspace{-2mm}
  \caption{Pseudo Code of Active Teacher}  
  \label[Algorithm]{alg:pseudocode_of_active_teacher}  
  \begin{algorithmic}[1]  
    \Require  
      Labeled Dataset \{$\mathcal{X}_L^0$, $\mathcal{Y}_L^0$\}, Unlabeled Dataset \{$\mathcal{X}_U^0$\}, Maximum Iteration $K$
    \Ensure  
      Teacher Model $M^t$
    \ForAll{$x_l \in \mathcal{X}_L^0$ and $x_u \in \mathcal{X}_U^0$}
        \State Update the parameters of Student $M^s_0$ by \cref{eq:semi-loss}
        \State Update the parameters of Teacher $M^t_0$ by \cref{eq:ema}
    \EndFor 
\ForAll{i=1,...,K} 
    \ForAll {$x_u \in \{\mathcal{X}_U^{i-1}\}$} 
        \State Calculate sampling score of unlabeled data using Teacher network $M_{i-1}^t$ by \cref{eq:lp-norm};
    \EndFor
    \State Rank the data based on score. 
    \State Select the top-N data \{$\mathcal{X}_P^i$\} and annotate them with label \{$\mathcal{Y}_P^i$\};
    \State Update labeled set \{$\mathcal{X}_L^i, \mathcal{Y}_L^i\} = \{\mathcal{X}_L^{i-1}, \mathcal{Y}_L^{i-1}\} \cup \{\mathcal{X}_P, \mathcal{Y}_P$\};
    \State Update unlabeled set \{$\mathcal{X}_U^i$\} = \{$\mathcal{X}_U^{i-1}$\} - \{$\mathcal{X}_P$\}
    \ForAll{$x_l \in \mathcal{X}_L^i$ and $x_u \in \mathcal{X}_U^i$}
        \State Update the parameters of Student $M^s_i$ by \cref{eq:semi-loss}
    \State Update the parameters of Teacher $M^t_i$ by \cref{eq:ema}
    \EndFor
\EndFor \\
\Return $M^t_K$
    \end{algorithmic}
\end{algorithm} 
% \vspace{+0.5mm}

\subsection{Semi-Supervised Learning}
\label[section]{subsec:semi-supervised}
Given a set of labeled data $\mathcal{D}_L=\{\mathcal{X}_L, \mathcal{Y}_L\}$ and a set of unlabeled data $\mathcal{D}_U=\{\mathcal{X}_U\}$, where $\mathcal{X}$ denotes the examples and $\mathcal{Y}$ is the label set, the target of semi-supervised learning is to maximize model performance based on both labeled and unlabeled data. 

Similar to prior works~\cite{tang2021humbleteacher-ssod-ts, unbiasedteacher-ssod-ts}, our semi-supervised learning paradigm also includes two detection networks with the same configurations, namely \emph{Teacher} and \emph{Student}, as shown in \cref{fig:framework}. In this paper, we use Faster-RCNN~\cite{fasterrcnn} as our baseline detection network. The teacher network is in charged of pseudo-label generation, while the student one is optimized with both ground-truth and pseudo-labels. Specifically, the optimization loss for the student network can be defined as:
\begin{equation}
    \mathcal{L} = \mathcal{L}_{sup} + \lambda \cdot \mathcal{L}_{unsup},
    \label[equation]{eq:semi-loss}
\end{equation}
where $\mathcal{L}_{sup}$ and $\mathcal{L}_{unsup}$ denote the losses for supervised and unsupervised learning, respectively, and $\lambda$ is the hyper-parameter to trade-off between $\mathcal{L}_{sup}$ and $\mathcal{L}_{unsup}$. 

For object detection,  $\mathcal{L}_{sup}$ consists of the classification loss $\mathcal{L}_{cls}$ of RPN and ROI head,  and the one for bounding box regression $\mathcal{L}_{loc}$. Then, $\mathcal{L}_{sup}$ is defined as
\begin{equation}
\label[equation]{equ:supervised-equ}
\begin{aligned}
    \mathcal{L}_{sup}=\frac{1}{N_l} \sum_{i=1}^{N_l}(\mathcal{L}_{cls}(x_l^i, y_{cls}^i) + \mathcal{L}_{loc}(x_l^i, y_{loc}^i)),
\end{aligned}
\end{equation}
where $\mathcal{L}_{cls}$ and $\mathcal{L}_{loc}$ are calculated by 
\begin{equation}
\begin{aligned}
\label[equation]{equ:cls-loc}
    &\mathcal{L}_{cls}(x_l^i, y_{cls}^i) = \mathcal{L}_{cls}^{rpn}(x_l^i, y_{cls}^i) + \mathcal{L}_{cls}^{roi}(x_l^i, y_{cls}^i), \\
    &\mathcal{L}_{loc}(x_l^i, y_{loc}^i) = \sum_{c \in \{\text{x,y,h,w}\}}\text{Smooth}_{L1}(t_c^{i} - y_c^i).
\end{aligned}
\end{equation}
Here, $x_l$ refers to the labeled example, $y_{cls}$ and $y_{loc}$ are its labels, and $N_{l}$ denotes the number of $x_l$. $t_c$ is the c-th coordinate of the output image $x_i$. In terms of $L_{loc}$, we use the smooth \emph{L}-1 loss for the bounding box regression:
\begin{equation}
    \text{Smooth}_{L1}(x)=\left\{
    \begin{aligned}
    &0.5x^2 &\text{if  } |x|<1,\\
    &|x|-0.5 &\text{otherwise}.
    \end{aligned}
    \right.
\end{equation}
For $\mathcal{L}_{unsup}$, we only use the pseudo-labels of RPN and ROI head predictions, similar to that in~\cite{unbiasedteacher-ssod-ts}. It is formulated as
\begin{equation}
    \mathcal{L}_{unsup}=\frac{1}{N_u} \sum_{i=1}^{N_u}\mathcal{L}_{cls}(x_u^i, \hat{y}_{cls}^i),
\end{equation} 
where $\mathcal{L}_{cls}$ is the same as \cref{equ:supervised-equ}, and $\hat{y}_{cls}^i$ is the pseudo-labels generated by the teacher network.

To avoid the class-imbalance and over-fitting issues, we follow~\cite{unbiasedteacher-ssod-ts, tang2021humbleteacher-ssod-ts} to freeze the optimization of the teacher network during semi-supervised training, and update its parameters from the student network via \emph{Exponential Moving Average} (EMA)~\cite{meanteacher-ssl-ts}:
\begin{equation}
    \theta_t^i \xleftarrow{} \alpha \theta_t^{i-1} + (1-\alpha)\theta_s^i,
    \label[equation]{eq:ema}
\end{equation}
where $\theta_t$ and $\theta_s$ are the parameters of the teacher and student networks, respectively, and $i$ denotes the $i$-th training step. $\alpha$ is the hyper-parameter to determine the speed of parameter transmission, which is normally close to 1. To improve the quality of pseudo-labels, we also apply non-maximum suppression (NMS)~\cite{rcnn} and confidence threshold to filter repetitive and uncertain pseudo-labels. 

\subsection{Active Sampling}
\label[subsec]{subsec:activesampling}
In Active Teacher, the label set is partially initialized and augmented through the teacher network after each semi-supervised training. We explore what kind of examples (or images) are critical for semi-supervised object detection, and introduce three active sampling metrics, namely \emph{difficulty}, \emph{information} and \emph{diversity}. 

\textbf{Difficulty} is the widely-used metric for active learning\cite{active-entropy, cho2021mcdal}, and is normally measured based on the entropy of the probability distribution predicted by the model. A higher entropy shows that the model is more uncertain about its prediction, suggesting that the example is more difficult. 

In SSOD, we measure the difficulty score $s_i^{\text{diff}}$ of an unlabeled example based on the category prediction of the teacher network, which is defined as
\begin{equation}
    s_i^{\text{diff}} = - \frac{1}{n_b^i} \sum_{j=1}^{n_b^i}  \sum_{k=1}^{N_c} p(c_k;b_j,\theta_t) \log p(c_k;b_j,\theta_t),
    \label[equation]{eq:difficult}
\end{equation}
where $n_b^i$ is the number of the predicted bounding box after NMS and confidence filtering, $N_c$ is the number of object categories and $p(c_k;b_j,\theta_t)$ is the prediction probability of the $k\text{-th}$ category by the teacher network. With \cref{eq:difficult}, we can judge whether the image is difficult for SSOD based on the prediction uncertainty of the teacher network.

\textbf{Information} is a metric to measure the amount of information of the unlabeled image for SSOD. In some classification tasks~\cite{cho2021mcdal, active-entropy}, it is often calculated by prediction entropies, similar to \emph{difficulty}. However, in object detection, richer information means that more visual concepts appear in the image, so the model can learn more detection patterns. To this end, we use the prediction confidence to measure this metric:
\begin{equation}
    s_i^{\text{info}} = \sum_{j=1}^{n_b^i} \text{confidence}(b_j,\theta),
    \label[equation]{eq:information}
\end{equation}
where the $\text{confidence}(b_j,\theta_t)$ is the  highest confidence score in $j\text{-th}$ bounding box predicted by the teacher network. From \cref{eq:information}, we can see that the larger $s^{\text{info}}$, the more visual concepts recognized by the teacher network, suggesting that the image has richer information. 

\textbf{Diversity} is a metric to measure the distribution of object categories in an image. The diversity score $s^{\text{dive}}$ is calculated by
\begin{equation}
    s_i^{\text{dive}} = |\{c_j\}_{j=1}^{n_b^i}|
\end{equation}
where $c_j$ is the predicted category of the $j-$th bounding box, and $|\cdot|$ is the cardinality. The difference between information and diversity is that the former will sample images of more visual instances that might belong to only one or a few categories, while the later will favor those involving more different concepts. 

\textbf{Metrics Combination.} The introduced metrics may be able to answer which type of examples are suitable for SSOD. However, a practical problem is that the models in different states may have different requirements for label information. Besides, how to maximize the benefits of these metrics without extensive trials remains a challenge. To this end, we propose a simple yet efficient solution to automatically combine these metrics, termed \emph{AutoNorm}.

Before combining these metrics, we notice that the value ranges of these metrics differ greatly. For instance, the \emph{difficulty} scores is usually between 0.3 and 0.8 with a theoretical maximum of $\log N_{c}$, while the \emph{information} score often ranges from 4.0 to 6.0. In this case, the first step of combination is to normalize their values:
\begin{equation}
    \hat{s_i^m} = \frac{s_i^{m}}{s_{\text{max}}^{m}}
\end{equation}
where $m \in \{\text{\emph{difficulty}}, \text{\emph{information}}, \text{\emph{diversity}}\}$ represent the metrics, the $s_{\text{max}}^{m}$ is the maximum value of this metric.

Since these metrics represent image information from different aspects, we further build a three-dimensional sampling space to represent each example as $\Vec{s_i} = (s_i^{\text{diff}}, s_i^{\text{info}}, s_i^{\text{dive}})$. The evaluation result of each unlabeled example can be regarded as a point in this space. Afterwards, we use \emph{L-p} normalize the data points into a single scalar $s_{L_p}$, which is obtained by

\begin{equation}
    s_{L_p} = L_p(\vec{s}) = ||\textbf{s}||_p = \sqrt[p]{\sum_{i=1}^3 s_i^p}
    \label[equation]{eq:lp-norm}
\end{equation}
where $\vec{s} = (s_1, s_2, s_3) = (\hat{s_i^{\text{diff}}}, \hat{s_i^{\text{info}}}, \hat{s_i^{\text{dive}}})$. Empirically, we use $L_1$ norm to combine these three metrics. When using $L$-$p$ (p\textgreater1) norm, the metrics with higher values will receive more sampling weights, \emph{e.g.}, \emph{difficulty}, which is found to be suboptimal in our experiments.

\begin{table*}[ht]
\caption{Comparison between the proposed Active Teacher and other SSOD methods on MS-COCO \emph{val2017}. The metric we used is mAP (50:95). “Supervised” refers to the performance of the model trained with labeled data only. * is our re-implementation. $\Delta$: AP gain to the supervised performance. Our method consistently outperforms the compared methods.}
\vspace{-2mm}
\setlength{\tabcolsep}{2mm}
\begin{tabular}{ccccccccccc}
\toprule
                 & \multicolumn{10}{c}{COCO-Standard}                                                            \\ \cline{2-11} 
                 & 1\%   & $\Delta$ & 2\%   & $\Delta$ & 5\%   & $\Delta$ & 10\%  & $\Delta$ & 20\%   & $\Delta$ \\ \hline
Supervised~\cite{fasterrcnn}       & 9.05  & +0.00     & 12.70 & +0.00     & 18.47 & +0.00     & 23.86 & +0.00     & 26.88* & +0.00*     \\ \hline
STAC~\cite{stac-ssod-ts}\mypm{arXiv2020}             & 13.97 & +4.92    & 18.25 & +5.55    & 24.38 & +5.91    & 28.64 & +4.78    & /      & /        \\
ISMT~\cite{ismt-ssod-ts}\mypm{CVPR2021}             & 18.88 & +9.83    & 22.43 & +9.73    & 26.37 & +7.90    & 30.53 & +6.67    & /      & /        \\
Instant-Teaching~\cite{instant-ssod-ts}\mypm{CVPR2021} & 18.05 & +9.00    & 22.45 & +9.75    & 26.75 & +8.28    & 30.40 & +6.54    & /      & /        \\
Humble-Teacher~\cite{tang2021humbleteacher-ssod-ts}\mypm{CVPR2021}   & 16.96 & +7.91    & 21.72 & +9.02    & 27.70 & +9.23    & 31.61 & +7.75    & /      & /        \\
Unbiased-Teacher~\cite{unbiasedteacher-ssod-ts}\mypm{ICLR2021} & 20.75 & +11.70   & 24.30 & +11.60   & 28.27 & +9.80     & 31.50 & +7.64    & 34.88* & +8.00*    \\ \hline
Active-Teacher(Ours) & \textbf{22.20} & \textbf{+13.15} & \textbf{24.99} & \textbf{+12.29} & \textbf{30.07} & \textbf{+11.60} & \textbf{32.58} & \textbf{+8.72} & \textbf{35.49} & \textbf{+8.61} \\ \bottomrule
\end{tabular}
\vspace{-3mm}
\label[table]{table: result}
\end{table*}

%%%%%%%%%%%%%%%%%% Experiment %%%%%%%%%%%%%%%%%%
\section{Experiment}
\subsection{Dataset and Metric}

We evaluate our approach on the main benchmark for object detection, namely MS-COCO~\cite{coco}. Specifically, MS-COCO divides the examples into two splits, namely \emph{train2017} and \emph{val2017}. The \emph{train2017} has 118k labeled images. During our experiments, this split is further divided into the labeled set and the unlabeled one, similar to the prior works in SSOD~\cite{stac-ssod-ts, unbiasedteacher-ssod-ts}. In practice, we adopt the settings of 1\%, 2\%, 5\%, 10\% and 20\% labeled data of \emph{train2017} for experiments and the comparisons with the other SSOD methods~\cite{stac-ssod-ts, ismt-ssod-ts, instant-ssod-ts, tang2021humbleteacher-ssod-ts, unbiasedteacher-ssod-ts}. The rest examples are regarded as unlabeled data. In terms of model evaluation, we follow the previous works~\cite{stac-ssod-ts, ismt-ssod-ts, instant-ssod-ts, tang2021humbleteacher-ssod-ts, unbiasedteacher-ssod-ts, consistency-based-ssod} adopt mAP (50:95)~\cite{coco} as the metric of our experiments. And \emph{val 2017}, which has 5k images, is used for evaluation. 

\subsection{Experimental Settings}
\label[section]{subsec:implementation}
Following the most work in SSOD~\cite{stac-ssod-ts, ismt-ssod-ts, instant-ssod-ts, tang2021humbleteacher-ssod-ts, unbiasedteacher-ssod-ts, consistency-based-ssod}, we use Faster-RCNN with ResNet-50 as our baseline detection network. The implementation and hyper-parameter setting are the same as those in Detectron2~\cite{dt2}. In terms of semi-supervised learning, we also follow the works in~\cite{unbiasedteacher-ssod-ts} to pre-train the teacher network with the supervised objectives defined in \cref{equ:supervised-equ}. The numbers of pre-training steps is set to 2k for all experimental settings. Afterwards, the student network is initialized with the parameters of the teacher one. The total training steps for each semi-supervised learning are 180k. The optimizer used is SGD~\cite{sgd}, and the learning rate linearly increases from 0.001 to 0.01 at the first 1k iterations, and is divided by 10 at 179,990 iteration and 179,995 iteration, respectively. Similar to~\cite{unbiasedteacher-ssod-ts}, we apply \emph{random horizontal flip} as weak augmentation for the teacher, and the strong augmentations for the student include \emph{horizontal flip}, \emph{color jittering}, \emph{grey scale}, \emph{gaussian blur} and \emph{CutOut}~\cite{cutout}. We use a threshold $\tau=0.7$ to filter the pseudo-labels of low quality. We use $\alpha=0.9996$ for EMA and $\lambda=4$ for the unsupervised loss on all experiments. In terms of active sampling, we set the iteration number in \cref{alg:pseudocode_of_active_teacher} as 2 in this paper. For all experiments, half of the label set are randomly selected, and the other half are actively selected after semi-supervised learning. The batch size is set to 64, which consists 32 labeled and 32 unlabeled images via random sampling.

% VOC part
% Experiments setting on \emph{VOC} is the same as above, except  \emph{VOC} randomly sample 5\% of \emph{VOC07 trainval} and \emph{VOC12 trainval} and active sampling another 20\%.

\subsection{Experimental Result}
\subsubsection{Quantitative Comparisons}
\textbf{Comparisons with the state-of-the-arts.} We first compare Active Teacher with a set of teacher-student based SSOD methods, of which results are given in \cref{table: result}. From this table, we can first observe that all teacher-student based methods greatly outperform the traditional supervised learning. Besides, we can also notice that with the careful designs in framework, those recently proposed teacher-student methods, \emph{e.g.} Unbiased Teacher~\cite{unbiasedteacher-ssod-ts}, improve the pioneer obviously, \emph{i.e.} STAC, suggesting the notable progresses made in teacher-student based SSOD. However, their competition also becomes more fierce. Even so, we still observe that the proposed Active Teacher can achieve obvious performance gains on all experimental settings, \emph{e.g.}, +6.3\% than Unbiased-Teacher with 5\% label information. These results greatly confirm the effectiveness of our method. 

\begin{table}[t]
\caption{Experiment of how much labeled data is for achieve 100\% supervised performance(37.63\cite{unbiasedteacher-ssod-ts}) by Unbiased-Teacher~\cite{unbiasedteacher-ssod-ts} and our Active-Teacher on MS-COCO.}
\label[table]{table:tosupervision}
\vspace{-3mm}
\centering
\begin{tabular}{ccccc}
\toprule
                 & \multicolumn{4}{c}{COCO-Standard} \\ \cline{2-5} 
                 & 5\%     & 10\%    & 20\%   & 40\% \\ \hline
Unbiased-Teacher & 28.27   & 31.50   & 34.88 &  37.29   \\
Active-Teacher   & 30.07 & 32.58 & 35.49    &  37.92    \\ \bottomrule
\end{tabular}
\end{table}

\begin{table}[t]
\caption{The result of Active Teacher on STAC~\cite{stac-ssod-ts}. We just replace the initial data while keep the rest settings the same.}
\vspace{-2mm}
\centering
\setlength{\tabcolsep}{4mm}
\begin{tabular}{cccc}
\toprule
                    & \multicolumn{3}{c}{COCO-Standard}                   \\ \cline{2-4} 
                    & 1\% & 5\% & 10\% \\ \hline
STAC                &  13.97   &   24.38   &  28.64\\
STAC+Ours           &   14.79   &   26.19    &   29.77   \\ \bottomrule
\end{tabular}
\vspace{-5mm}
\label[table]{table:stac+ours}
\end{table}

\textbf{Requirement of labeled data to achieve supervision.}
In practical applications, the minimum amount of labeled data required to achieve supervised performance is more concerned. For this purpose, we conduct a comparison between Unbiased-Teacher\cite{unbiasedteacher-ssod-ts} and our Active Teacher. As shown in \cref{table:tosupervision}, with 40\% labeled data our method could achieve supervised performance easily.

\begin{figure}[t]
    \centering
    \includegraphics[width=1\columnwidth]{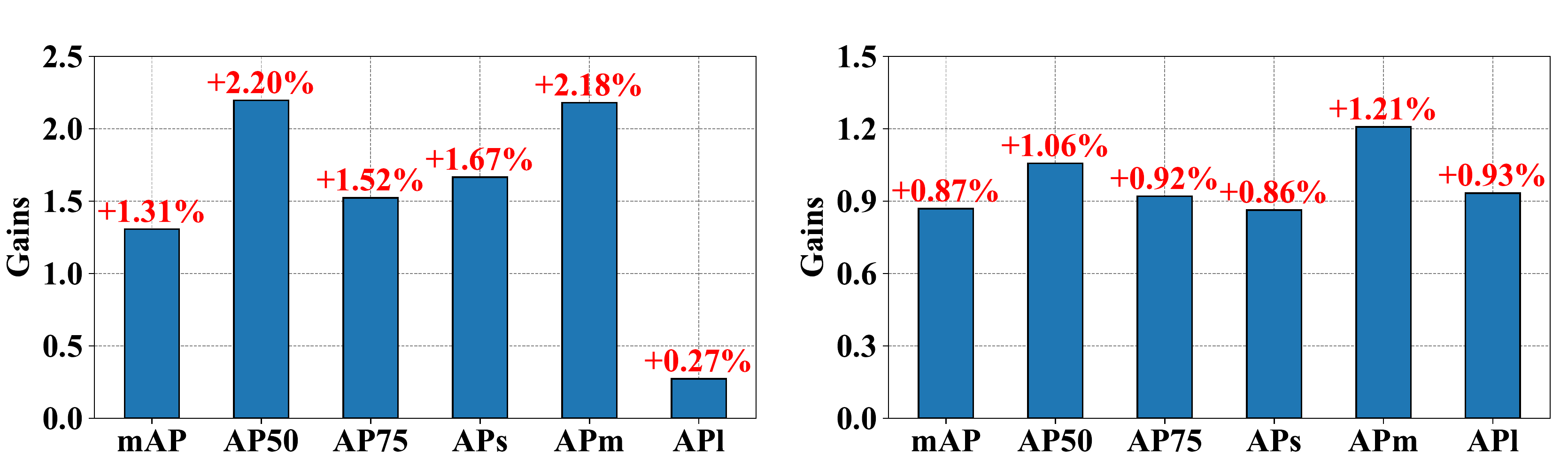}
    \vspace{-4mm}
    \caption{Changes of specific AP indicators of Active Teacher compared with  Unbiased Teacher on 5\% and 20\% labeled data. Active Teacher is more sensitive to small and medium sized object.}
    \vspace{-2mm}
    \label[fig]{fig:detail_mAP}
\end{figure}

\textbf{Effect of Active Teacher on different AP metrics.} \cref{fig:detail_mAP} shows the detailed performance gains of Active Teacher against Unbiased Teacher on more metrics. On 5\% labeled data, Active Teacher can greatly improve the performance on the detection of medium and small objects, \emph{i.e.}, APs and APm, suggesting that Active Teacher can sample images with more small objects. On 20\% labeled data, all AP metrics can obtain obvious improvements by Active Teacher, which also suggests its change in data sampling.

\textbf{Generalization capability of active sampling.} Active Teacher is also highly generalized. \cref{table:stac+ours} illustrates the performance changes of STAC after using the selected label information by Active Teacher. Without bells and whistles, this simple modification can lead to obvious performance gains of STAC on all experimental settings, strongly suggesting the generalization ability of our method.

\begin{figure}[t]
    \centering
    \includegraphics[width=1\columnwidth]{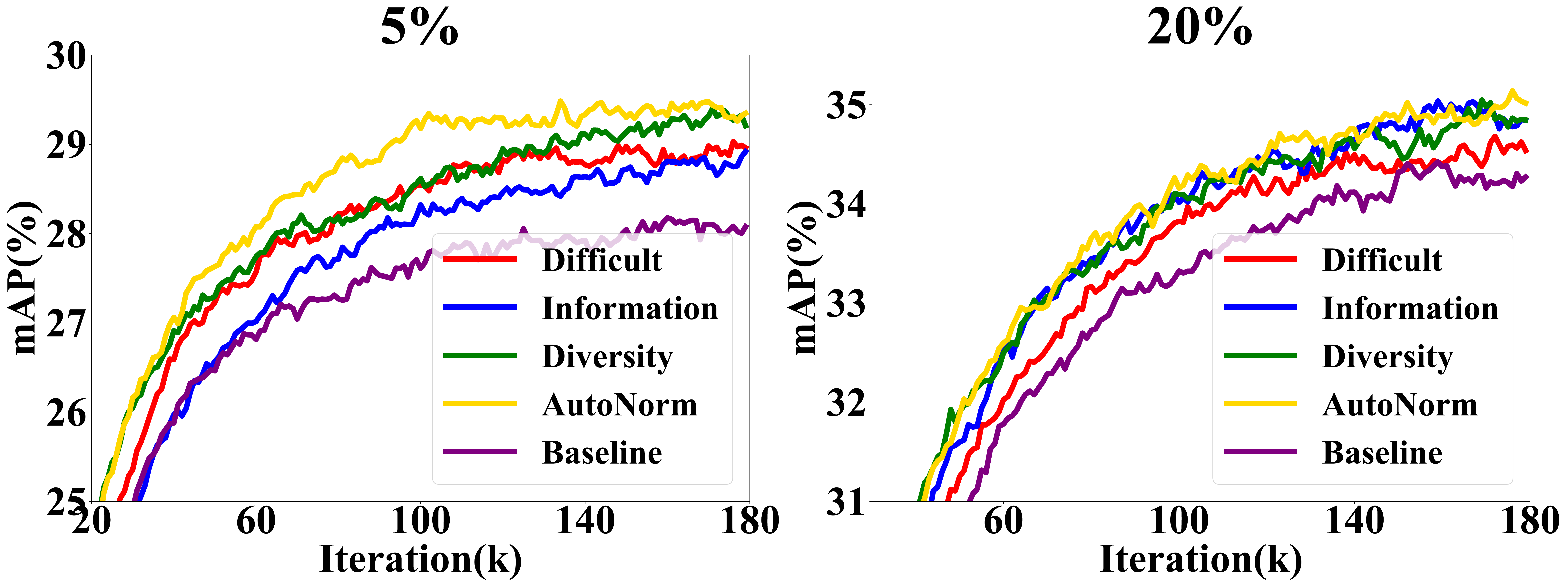}
    \vspace{-6mm}
    \caption{Training curves of active sampling with different sampling metrics on 5\% and 20\% labeled data. The proposed AutoNorm can well combine the advantages of three metrics.}
    \vspace{-1mm}
    \label[fig]{fig:curve}
\end{figure}

% \usepackage{multirow}
\begin{table*}[htbp]
\caption{Ablation study of different sampling strategies in Active Teacher. Note that these results are experimented with a smaller batch size, \emph{i.e.} 32, which are slightly inferior than those in \cref{table: result}.}
\vspace{-3mm}
\centering
\begin{tabular}{ccccccc}
\toprule
\multirow{2}{*}{Strategy} & \multicolumn{3}{c}{Metric}   & \multicolumn{3}{c}{COCO-Standard}                                                          \\ \cline{2-7} 
& Difficulty & Information & Diversity & 5\%(2.5\%+2.5\%) & \multicolumn{1}{l}{10\%(5\%+5\%)} & \multicolumn{1}{c}{20\%(10\%+10\%)} \\ \hline
Baseline & - & -  & -  & 27.84 & 31.39 & 34.26*    \\ \hline
Difficulty  & \checkmark & -  & -  & 29.03  & 32.13  & 34.68  \\
Information & -  & \checkmark & -  & 28.92 & 31.98 & 35.04 \\
Diversity   & -  & -  & \checkmark &  29.40 & 32.26 & 35.05 \\
AutoNorm    & \checkmark & \checkmark & \checkmark & 29.48 & 32.08 & 35.13 \\ \bottomrule
\end{tabular}
\vspace{-2mm}
\label[table]{table:ablation}
\end{table*}

\begin{figure*}[ht]
    \centering
    \begin{minipage}[c]{0.96\textwidth}
        \centering
        \includegraphics[width=\textwidth]{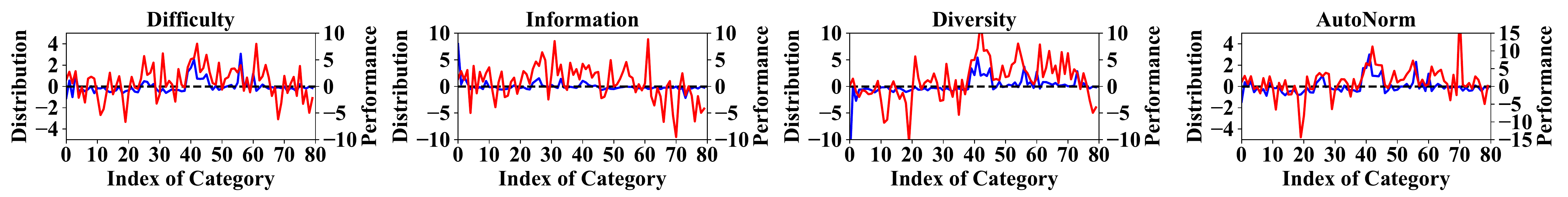}
        \vspace{-5mm}
        \subcaption{Results on 5\% labeled data.}
        \label[subfig]{fig:distribution2.5+2.5}
    \end{minipage} \\

    \begin{minipage}[c]{0.96\textwidth}
        \centering
        \includegraphics[width=\textwidth]{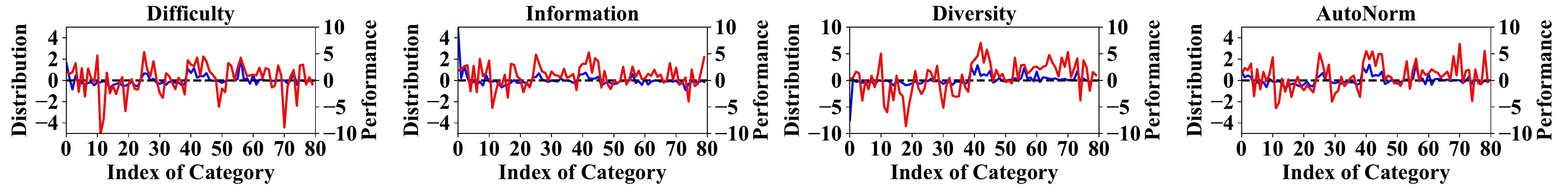}
        \vspace{-5mm}
        \subcaption{Results on 20\% labeled data.}
        \label[subfig]{fig:distribution10+10}
    \end{minipage} \\
    \vspace{-3mm}
    \caption{The relative changes of sample distribution (\textbf{blue})  and performance (\textbf{red}) of Activate Teacher with and without active sampling on different metrics. The results are obtained on 5\% and 20\% labeled data.}
    \vspace{-5mm}
    \label[fig]{fig:distribution0category}
\end{figure*}

\textbf{Ablation.} We also ablate the proposed metrics with different proportions of labeled data, as shown in \cref{table:ablation}. From this table, we can see that three metrics, \emph{i.e.}, \emph{difficulty}, \emph{information} and \emph{diversity}, are all beneficial for SSOD. However, under different settings of label proportions, their performance is also different, which verifies the assumption we made in \cref{subsec:activesampling}. For instance, with more label examples, the metric of information will performs better, and \emph{vice verse}. In addition, as shown in \cref{fig:curve}, AutoNorm is superior than the other metrics during the training and obtains the overall better performance finally, which well confirms its effectiveness.

\textbf{Sampling distributions and performance changes.}
To obtain deep insight into these metrics, we further compare their detailed sampling distributions and performance changes on all categories, of which results are depicted in \cref{fig:distribution0category}. From these results, we can first observe that \emph{information} is easy to suffer from  \emph{inverse compensation effect}. Specifically, the categories that already take a large proportion of data will receive more samplings from this metric. As a result, the biased distribution and unbalanced performance will become more prominent under this metric. Notably, \emph{diversity} is the opposite of \emph{information}, which can also address \emph{diminishing marginal effect}. From \cref{fig:distribution0category}, we can find that the performance gains of the major categories will not keep increasing with more examples. In contrast, some small categories will obtain more improvements via data augmentation, which can be achieved by \emph{diversity}. However, due to the obvious difference between its sampling distribution and the real one, the advantage of \emph{diversity} will be weaken as the number of labeled examples increases. The distribution of \emph{difficulty} matches the real one. Due to the preference of outliers, its overall performance is not significant. Instead, the proposed \emph{AutoNorm} can make good use of three metrics, while maintaining the amount of information and the diversity of examples. Besides, it is also closer to the real data distribution.

\begin{figure*}[ht]
    \vspace{-5mm}
    \centering
    \begin{minipage}[c]{1.0\textwidth}
        \centering
        \includegraphics[width=\textwidth,height=0.2\textwidth, trim=10 180 10 180,clip]{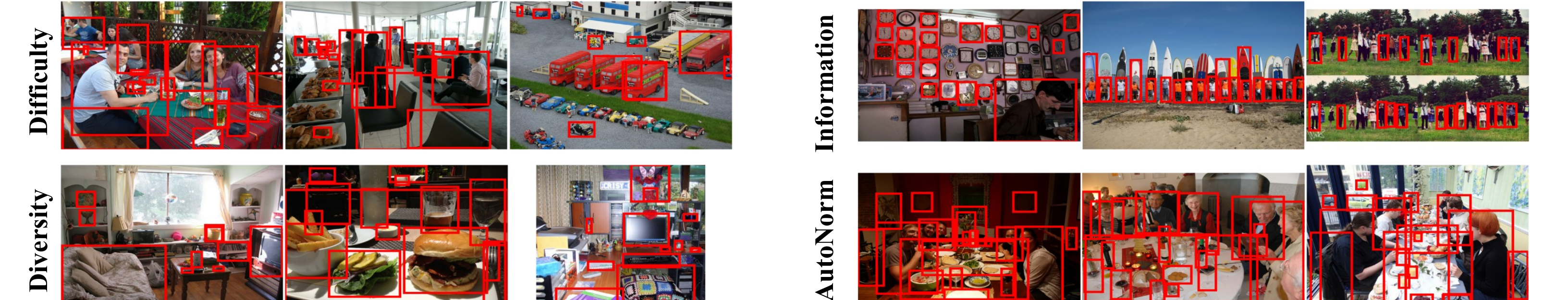}
        \subcaption{Images selected by different metrics with 5\% labeled data.}
        \label[subfig]{fig:visualization_metric5}
    \end{minipage} \\
    
    \begin{minipage}[c]{1.0\textwidth}
        \centering
        \includegraphics[width=\textwidth,height=0.2\textwidth, trim=10 180 10 170,clip]{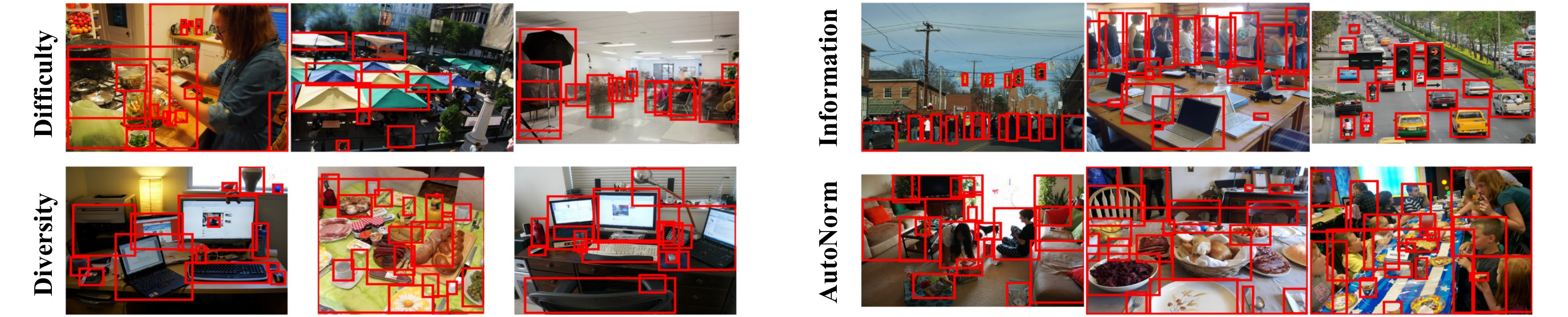} 
        \subcaption{Images selected by different metrics with 20\% labeled data.}
        \label[subfig]{fig:visualization_metric20}
    \end{minipage} \\
    \vspace{-3mm}
    \caption{Visualization of the images with top ranks with 5\% and 20\% labeled proportions and different sampling metrics. The bounding boxes in red are predicted by the teacher network.}
    \vspace{-1mm}
    \label[fig]{fig:visualization}
\end{figure*}

\begin{figure*}[ht]
    \vspace{-1mm}
    \centering
    \includegraphics[width=1.0\textwidth,trim=0 120 0 130,clip]{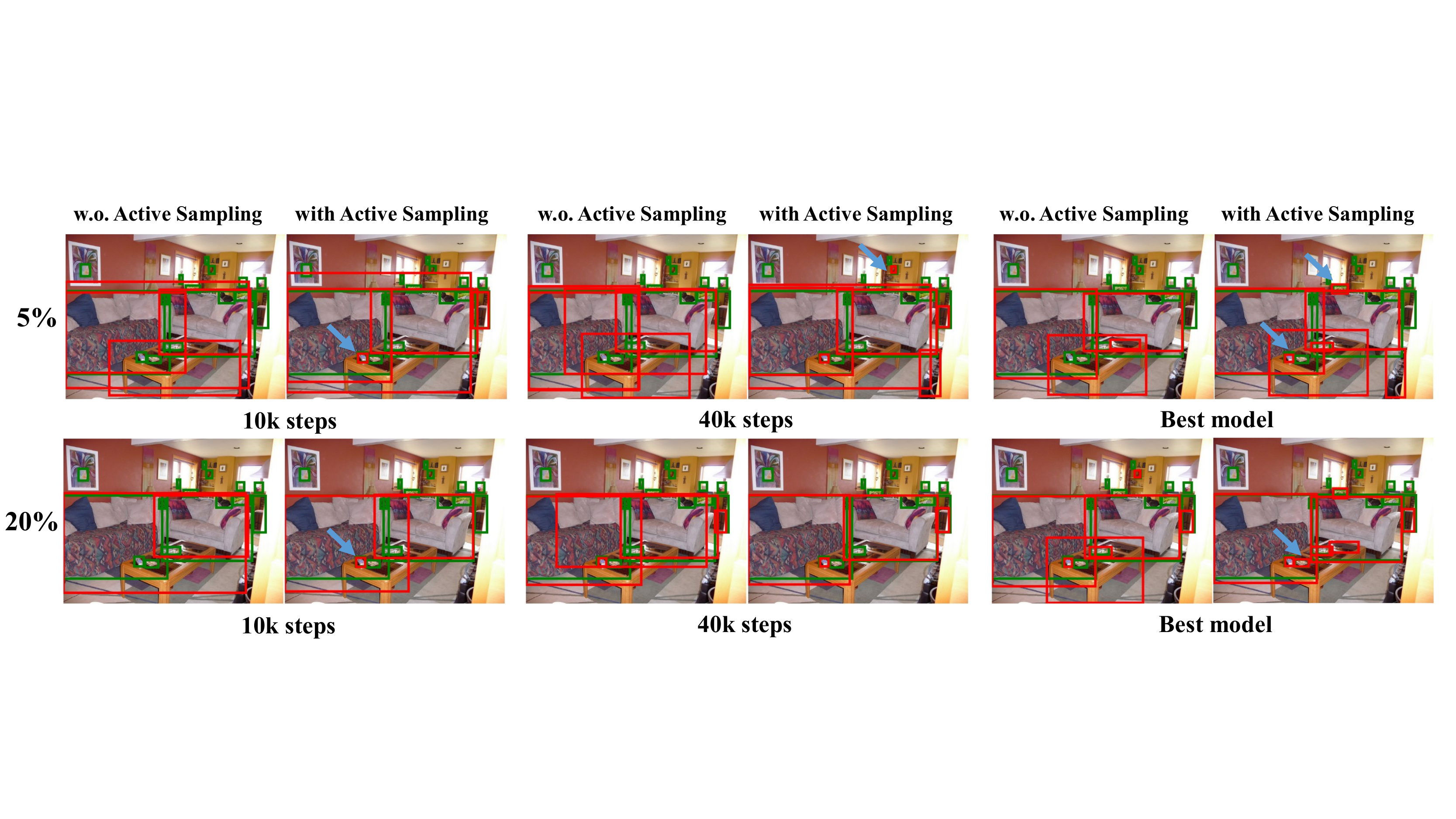}
    \vspace{-6mm}
    \caption{Visualization of the pseudo-labels predicted by Active Teacher with and without active sampling at different training steps. The \textbf{green} bounding boxes are the ground-truths, while the \textbf{red} ones are pseudo-labels predicted by the teacher network.}
    \vspace{-3mm}
    \label[fig]{fig:vis_pseudo_different_iteration}
\end{figure*}

% \vspace{-4mm}
\subsubsection{Qualitative analysis}
\vspace{-2mm}

\textbf{What examples are selected by these metrics?}
In \cref{fig:visualization}, we visualize the examples selected by these metrics based on 5\% and 20\% labeled data. From \cref{fig:visualization}, we can first observe that the selected examples well correspond to the definitions of these metrics. For instance, \emph{difficulty} will sample examples with objects that are difficult to detect, \emph{e.g.}, small objects, and \emph{information} prefers the ones with more instances, \emph{e.g.} street views. \emph{Diversity} will select the images containing more categories, \emph{e.g.} dining room. In addition, we can also notice some slightly difference between the samplings with 5\% and 20\% labeled data. Specifically, under 5\%, the teacher is not sufficiently trained, so it can only estimate the examples of the common categories. For instance, \emph{information} will sample a picture of only people, which also explains why its sampling is less effective on 5\%. In contrast, under 20\%, the example estimation becomes more comprehensive. Besides, we can find that the proposed AutoNorm is the optimal strategy on both settings. The images sampled by AutoNorm are \emph{full of information, rich in categories and different in object sizes}. We believe that this is also the proper criteria of data sampling for SSOD from an overall perspective. 

\textbf{Effects on pseudo-labels}. 
We further visualize the pseudo-labels of Active Teacher with and without active sampling on different training steps. Firstly, we can find that there is still an obvious gap between the qualities of the pseudo-labels and the ground-truth ones. Even so, with the help of active sampling, Active Teacher can still generate more pseudo-labels with better qualities in different training steps. As shown in \cref{fig:vis_pseudo_different_iteration}, Active Teacher can also detect more small objects in image. This result greatly confirms our argument that data initialization also affects the qualities of pseudo-labels.  

% \vspace{-3mm}
\section{Conclusion}
% \vspace{-2mm}
In this paper, we propose a novel teacher-student based method for semi-supervised object detection (SSOD), termed \emph{Active Teacher}. Different from prior works, Active Teacher studies SSOD from the perspective of data initialization, which is supported  with a novel active sampling strategy. Meanwhile, we also investigate the selection of examples from the aspects of \emph{information}, \emph{diversity} and \emph{difficulty}. The experimental results not only show the superior performance gains of Active Teacher over the existing methods, but also show that it can help the baseline network achieve 100\% supervised performance with much less label expenditure. Meanwhile, the quantitative and qualitative  analyses provide useful hints for the data annotation in practical applications.

\vspace{-1mm}
\noindent\textbf{Limitation.} A potential issue of Active Teacher is that it theoretically takes $k-1$ times more training steps than the other teacher-student methods, where k is the number of training iterations in \cref{alg:pseudocode_of_active_teacher}. In our experiments, we find that $k=2$ can already help the model obtain obvious performance gains. Considering the fact that data annotation is much more expensive than model training in some practical applications of SSOD, \emph{e.g.}, security surveillance and industrial inspection, we believe that the doubled training time is still acceptable.

\vspace{-1mm}
\noindent\textbf{\small Acknowledgements.} 
{\small
This work was supported by the National Science Fund for Distinguished Young Scholars (No.62025603), the National Natural Science Foundation of China (No.U21B2037, No.62176222, No.62176223, No.62176226, No.62072386, No.62072387, No.62072389, and No.62002305), Guangdong Basic and Applied Basic Research Foundation (No.2019B1515120049), and the Natural Science Foundation of Fujian Province of China (No.2021J01002). We also thank Huawei Ascend Enabling Laboratory for the continuous support in this work.}

%%%% References
{\small
\bibliographystyle{ieee_fullname}
\bibliography{ref.bib}
}

\end{document}